\def\gmode{} 

 \def\gdriver{dvipdfmx}

 \makeatletter 
 \@ifl@t@r{2014/06/01}\fmtversion{\def\gdriver{dviout}}{}

\def\my@if@primitive#1{%
  \edef\my@tmpa{\string#1}\edef\my@tmpb{\meaning#1}%
  \ifx\my@tmpa\my@tmpb \expandafter\@firstoftwo
  \else \expandafter \@secondoftwo \fi}
\my@if@primitive\luatexversion{\def\gdriver{}}{}

\newcommand{\dtitle}[1]{\title{ \if \gmode \else
\color{red} Demo mode!\\
comment out \textbackslash def \textbackslash gmode\{demo\} at the header to include figures \color{black}\\
\fi
#1 }}
\ifx\pdfoutput\undefined
\else
 \def\gdriver{}
\fi

\documentclass[\gmode,\gdriver]{article}
\usepackage{spconf,amsmath,graphicx}

\title{A method for adding motion-blur on arbitrary objects by using\\  
auto-segmentation and color compensation techniques}
\twoauthors
 {Michihiro Mikamo, Ryo Furukawa}
	{Hiroshima City University, Hiroshima, Japan}
 {Hiroshi Kawasaki}
	{Kyushu University, Fukuoka, Japan}

\graphicspath{{../}}

\usepackage{siunitx} 
\usepackage{color}

\ifx\argmin\undefined\newcommand{\argmin}{\mathop{\rm argmin}\limits}\fi
\ifx\argmax\undefined\newcommand{\argmax}{\mathop{\rm argmax}\limits}\fi
\ifx\bm\undefined\newcommand{\bm}[1]{\mbox{\boldmath{$#1$}}}\fi
\ifx\um\undefined\newcommand{\um}[1]{{\SI{#1}{\micro \metre}}}\fi 

\ifx\etal\undefined\newcommand{\etal}{{\it et al. }}\fi
\ifx\ie\undefined\newcommand{\ie}{{\it i.e.}}\fi
\ifx\eg\undefined\newcommand{\eg}{{\it e.g.}}\fi
\ifx\gt\undefined\newcommand{\gt}{$\textgreater$}\fi
\ifx\lt\undefined\newcommand{\lt}{$textless$}\fi

\ifx\figref\undefined\newcommand{\figref}[1]{{Fig.\ref{#1}}}\fi
\ifx\tabref\undefined\newcommand{\tabref}[1]{{Tab.\ref{#1}}}\fi
\ifx\equref\undefined\newcommand{\equref}[1]{Eq.(\ref{#1})}\fi
\ifx\secref\undefined\newcommand{\secref}[1]{Sec.\ref{sec:#1}}\fi
\ifx\subsecref\undefined\newcommand{\subsecref}[1]{Sec.\ref{sec:#1}}\fi

\begin{document}
%
\maketitle
\begin{abstract}
When dynamic objects are captured by a camera, motion blur inevitably occurs.
Such a blur is sometimes considered as just a noise, however, it sometimes 
    gives an important 
    effect to add dynamism in the scene for photographs or videos.
Unlike the similar effects, such as defocus blur, which is now easily  
    controlled even by smartphones, motion blur is still 
    uncontrollable and makes undesired effects on photographs.
In this paper, an unified framework to add motion blur on per-object basis is proposed. 
In the method, multiple frames are captured without motion blur and they are 
    accumulated to create motion blur on target objects.
To capture images without motion blur, shutter speed must be short, 
    however, it makes captured images dark, and thus, 
a sensor gain should be increased to compensate it. 
Since a sensor gain causes a severe noise on image, 
we propose a color compensation algorithm based on 
    non-linear filtering technique for solution. 
Another contribution is that our technique can 
    be used to make HDR images for fast moving objects by using 
multi-exposure images. 
In the experiments, effectiveness of the method is confirmed by ablation study 
    using several data sets.

\end{abstract}
\section{Introduction} 
Recently, smartphones are usually equipped with high quality cameras and 
demands on special effects or eye-catching filters are increasing.
Among those special effects and filters, blur control is a basic function and 
installed in many cameras. For those cameras, only defocus blur can be  
controlled, but 
motion blur is not considered. One reason is that the motion blur 
is theoretically more difficult to control, 
since it requires not only static information, but 
also dynamic information. 
In this paper, we propose a method to add motion blur to arbitrary objects in 
the scene.
The uniqueness of the proposed method is that we solve the problem by using two contradictory images as input, that is, the one with noise but no motion-blur, and the other with no noise but having blur.
In our method, multiple frames are captured without motion blur and they are 
    accumulated to synthesize motion blur on target objects.
To capture input images without motion blur, we set shutter speed extremely short, 
    however, it makes captured images dark, and thus, a sensor 
    gain is increased to compensate it. Since it causes severe noises on 
    captured images, color compensation algorithm based on 
    non-linear filtering technique is proposed in the paper. 
    Then, to achieve adding motion blur on arbitrary objects in the captured 
    image, auto-segmentation method is applied.
    based on 
    Note that our segmentation 
    algorithm is robust, since images are blur-free by our color compensation 
    method.

Another contribution our method is that HDR image synthesis is achieved for fast moving objects by
capturing multi-exposure images.
To keep shutter speed short and constant, only sensor 
    gains are allowed to be changed frame by frame, which is easily achieved by existing 
    camera control softwares~\cite{magiclantern}. At the accumulation process, 
    common HDR-image synthesis algorithm is applied to each segment independently.

In the experiments, several videos are downloaded  from Internet and used to create 
arbitrary motion-blur images to show the 
effectiveness of our method. In addition, real images with multiple 
exposures 
are captured by using camera control 
software and HDR images with motion blur on arbitrary objects were synthesized. 
Comparison with previous method as well as  
ablation studies were conducted to confirm the effectiveness of our method.

\section{Related work}
Recently, there have been several attempts to synthesize motion blur efficiently. 
These techniques are mainly categorized in two. 
One approach is to capture blur-free background images 
by digital stabilization 
techniques~\cite{jeong2015digital,Navarro11, Wang13, Grundmann12}. 
The technique requires special high fps cameras and is only applicable to mostly static scenes.
The other approach is to capture sharp images and 
add blur to the background afterward using multiple
images
~\cite{lancelle2019controlling,2001-Brostow-IMBSMA,Telleen07,6977984,10.1145/2159616.2159639} 
or video~\cite{Liu14}.
These techniques are based on object tracking to extract foreground objects
and add blur to background.
One limitation is that they assume high quality images are used as input, 
however, this is not always true, especially when the object is moving fast.

\begin{figure*}
\centering
\includegraphics[width=0.85\linewidth, bb= 0 0 874 285]{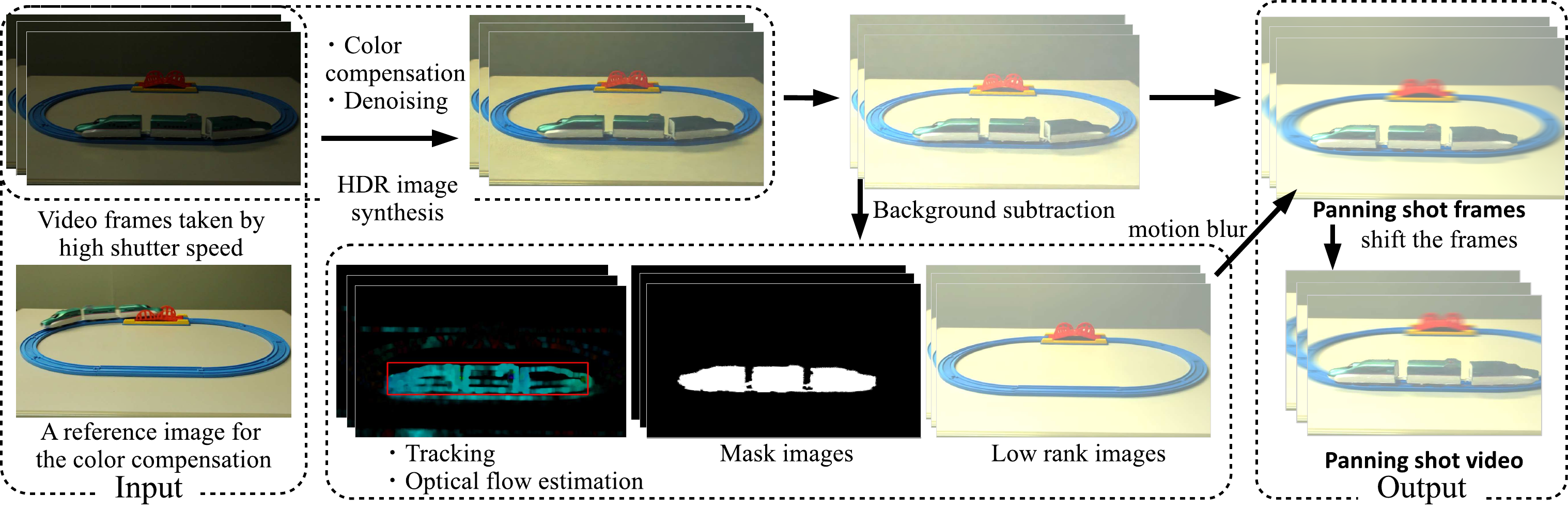}
\caption{Overview of the proposed method. Our method accepts a video taken at high shutter speed and a photo taken at proper exposure as input. 
The output is selective motion-blur added video in which the target remains sharp while 
    the other objects include motion blur.}
\vspace{-2mm}
\label{fig:overview}
\end{figure*}

\begin{figure}
\centering
\vspace{-1mm}
\includegraphics[width=0.90\linewidth, bb= 0 0 380 169]{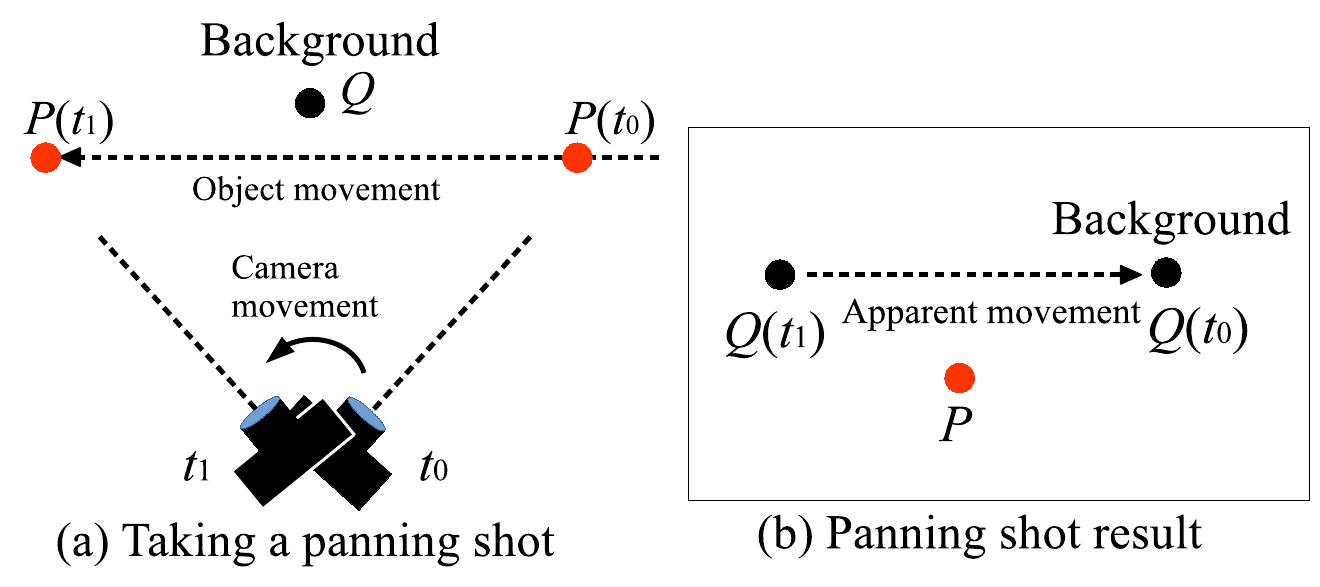}
\caption{The process of taking a panning shot. In (a), the user rotates the camera to follow the target object while the camera keeps its shutter open. By keeping the target on the same position, the result obtains the panning shot effect that the target remains in the same position while the background moves, shown in (b).
}
\vspace{-3mm}
\label{fig:PanningshotBackground}
\end{figure}

\vspace{-3mm}
\section{Overview}
\vspace{-3mm}
\label{sec:Overview}

\subsection{The algorithm}

To add motion-blur effects into 
movies
our idea is to 
capture a non-blurry image at fast shutter speed and compensate for the low quality due to under-exposure by using a high quality image captured at ordinary 
shutter speed.

An overview of the proposed method is shown in Fig.~\ref{fig:overview}. 
For the input, to get the blur-less images, the easiest way to capture such a 
videos is by using an ordinary video camera while setting the exposure time to be 
short.
However, setting such a short exposure time normally results in underexposed 
images, \ie, dark pixels with high noise. 
To cope with the problem, we use a color compensation method, where
we also capture the same scene with proper exposure as a 
reference image for compensation. 
This also helps to improve color consistency and visual smoothness of the resulting images~\cite{Reinhard01, Pitie07, Li18, He19}. Then, HDR image synthesis is possible using the images before and after color compensation is applied by regarding them as the multi-exposure image sets~\cite{Mann95, Debevec97, Nayar99, Reinhard10}.

Next, since the blurry effect is only required for the target object, images are separated 
into multiple segments based on object motion. 
To extract the target regions from the image sequence,  the area 
is designated by a bounding box in the first frame and tracked through the entire frame sequence.
The motion of the target region is then estimated by calculating optical flow of the bounding box.

Finally, 
motion blur is applied only to the background using the direction of optical 
flow, and the target object is added by an alpha blending technique using image masks.

\subsection{Examples of effects}

An example of effects that can be synthesized
is motion-blur for objects in movies.
In this effect, object motion can be analyzed by the optical flows.
Then the motion-blur effects can be added to the objects
with arbitrary lengths.
Normally, the lengths of the motion-blur effects is set to be proportional 
to the optical flow. 
In this case, the user can emphasize or suppress the motion-blur effects.

Another example of the effects that can be added
is a panning-shot, 
where the photograph is captured while the camera is tracking the target
with a long exposure time. 
This shot is known to be difficult even for professional photographers. 
Fig.~\ref{fig:PanningshotBackground} shows how to take a panning shot.
As shown in Fig.~\ref{fig:PanningshotBackground}(a), the camera follows the target $P$.
If the camera tracks the target perfectly, 
the target object is not affected by motion burr, while
the background is motion-blurred. 
To achieve this effect, only the background layer is blurred by the motion flow, 
which can be calculated from the optical-flow analysis. 

Aside from motion-blur effects, 
multi-frame operations other than motion-blur is also possible. 
In this paper, we propose high-dynamic-range (HDR) image synthesis.
HDR images can be synthesised from multiple images with different exposure times.
The proposed method can reconstruct multiple exposures of the same scene by using color transfer. 
By synthesizing HDR images from these multiple exposures, 
it is possible to achieve HDR frames that retain details both in bright and dark areas. 
For some cameras, exposure times can be controlled frame-by-frame basis. 
Using such a camera, we can also capture HDR images from even dynamic scenes.

\vspace{-3mm}
\section{Implementation}
\vspace{-3mm}
\subsection{Capturing input video and reference image}
\vspace{-1mm}
As described in the overview, 
we assume that the input video is not affected by motion blur. 
As an easy way to achieve this,  
we propose to capture videos
using a normal video camera
with a short exposure time. 
In our experiments, 
we used an off-the-shelf DSLR camera ~\cite{EOS60D}.
With the acquisition of video, we also capture the scene with an exposure time that is sufficiently long. 
Since this image represents the natural color of the scene, 
we use this image as the color reference image 
to compensate for the underexposure of the video.

\vspace{-3mm}
\subsection{Color transfer for underexposure compensation}

\label{sec:colortrans}
One of the problems in capturing the input video
with a short exposure time 
is that the amount of light which reaches to the sensor is reduced, 
resulting in underexposed, dark, and low quality images.
The most naive solution for the underexposure 
is just simply multiplying a constant value to the pixel values, however,
this often results in high noise with unnatural colors. 

To deal with this problem, 
we compensate for underexposure
by transferring the scene colors 
using the color reference image. 
Colors of each frame are converted so that
the resulting image becomes consistent with the color reference image. 
In this paper,
we used a color transfer developed by Piti\'{e} and Kokaram \cite{Pitie07}.
The method is based on the theory for the Monge-Kantorovitch problem, that is, a linear optimal transport problem.
This method can achieve good results in several scenes and is faster than the recent deep learning methods.
Besides, by using the pair of frames before and after color compensation, we can synthesis HDR frames. The color information can be efficiently retrieved by the multi-frames.

\vspace{-2mm}
\subsection{Foreground and background separation}
Each image frame of the the input video is 
separated into layers of the foreground and background
so that motion blur can be added differently to each layers. 
For adding different blur effects to the foreground and background 
the background layers should have occluded regions near the 
borders between the foreground and background. 
To achieve this, 
we use low-rank approximation extract the background layer,
assuming that background layer is temporary static.

In the process, we pack the input image sequence into 
a 2D matrix ${\bf A}$, where
a row represents an image frame
and a column represents a temporal intensity-value sequence of 
a single pixel.
Then, a low-rank approximation of ${\bf A}$ is 
calculated through the singular value decomposition
as follows:
\vspace{-1mm}
\begin{equation}
{\bf A = U \Sigma V^T},
\end{equation}
where ${\bf U}$ and ${\bf V}$ are 
the orthogonal matrices, ${\bf \Sigma}$ is a
rectangular diagonal matrix having singular values as diagonal components.
Then, the intensity signals of the background layer are extracted
as the first component, 
\ie, ${\bf A_{1}}$.
The second and the higher order components are dynamic layer signals. 
We used the method of Shu \etal~\cite{Shu14}, where 
low-rank images can be calculated with low computational costs.

\begin{figure}[tb]
\centering
\includegraphics[width=0.95\linewidth, bb= 0 0 520 446]{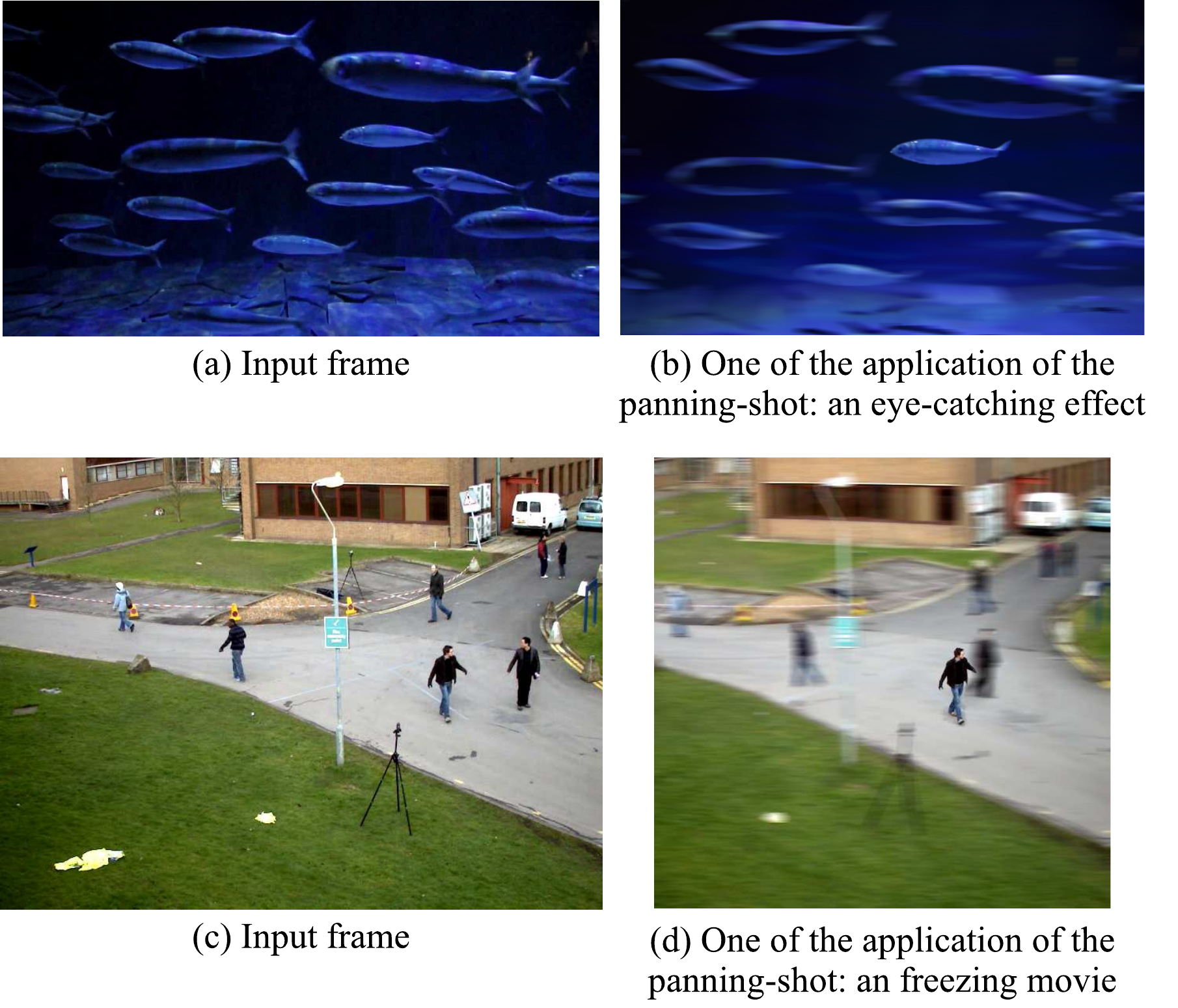}
\caption{Two examples of our method
synthesizing motion-blur on a fish and a human.
It represents the panning shot and freezing effects by stabilize the only selected object and 
    blur-out the others.
The original videos are downloaded from websites~\cite{fishVideo,Farneback03}.}
\label{fig:application}
\vspace{-2mm}
\end{figure}

\subsection{Synthesis of motion blur effect}

In the proposed method, 
the effect of motion blur 
is applied to the background layer. 
The amount and direction of motion blur 
is determined based 
on motion of the foreground. 

To estimate the foreground object motion, 
optical flow in the
foreground region is calculated.
The region of the target is specified by the user.
This allows the system to obtain the average motion of the 
foreground region,
which is used as the motion of the foreground object.
The method proposed by Farneback \etal\cite{Farneback03} was used to calculate the optical flow.

The motion blur applied to the background layer
is the inverse motion vector of the foreground,
which assumes that the position of 
the foreground region is fixed in the resulting movie. 
Fixation of the foreground region is not necessarily pixel-level.
For example, in many cases, 
the center of the bounding box is fixed.
In many applications, this is sufficient for emphasizing the dynamism.

An image layer ${\bf L}_e$ with a targeted motion-blur effect
can be computed by 
convolution from an input layer ${\bf L}_i$ and the kernel ${\bf K}$,
\ie, 
${\bf L}_e = {\bf K} \ast {\bf L}_i$.
${\bf K}$ is defined from the length and angle of the optical flows of the desired effect.
This can be either decided arbitrarily, or, decided from the average of the observed optical flows.

In case of HDR image effect, 
HDR image composition is processed from a multiple-exposed image set. 

Then, the modified layers
are integrated into a single output frame.
The layer integration is processed by 
weighted addition with masking operation.
For two layer cases with the foreground and background, we use 
\vspace{-2mm}
\begin{eqnarray}
{\bf I}_{f} = ({\bf 1} - {\bf M})\sigma{\bf L}_b + {\bf M}{\bf L}_e,
\end{eqnarray}
where ${\bf I}_{f}$ is the resulting image, 
${\bf M}$ is the mask image that obtained by, 
\eg, background subtraction method~\cite{Shu14}, 
and ${\bf L}_e$ is a layer modified by the desired effect. 
In the proposed method, the foreground and background are treated separately, 
therefore the relation of intensity between them tend to be lost. 
Here, we introduce $\sigma$ 
that retrieve the balance of the intensity between foreground and background, 
which is computed from before and after background subtraction.

\vspace{-2mm}
\section{Experiments}
\label{sec:Experiments}
\vspace{-2mm}
\begin{figure}[t]
\centering
\includegraphics[width=0.83\linewidth, bb= 0 0 375 297]{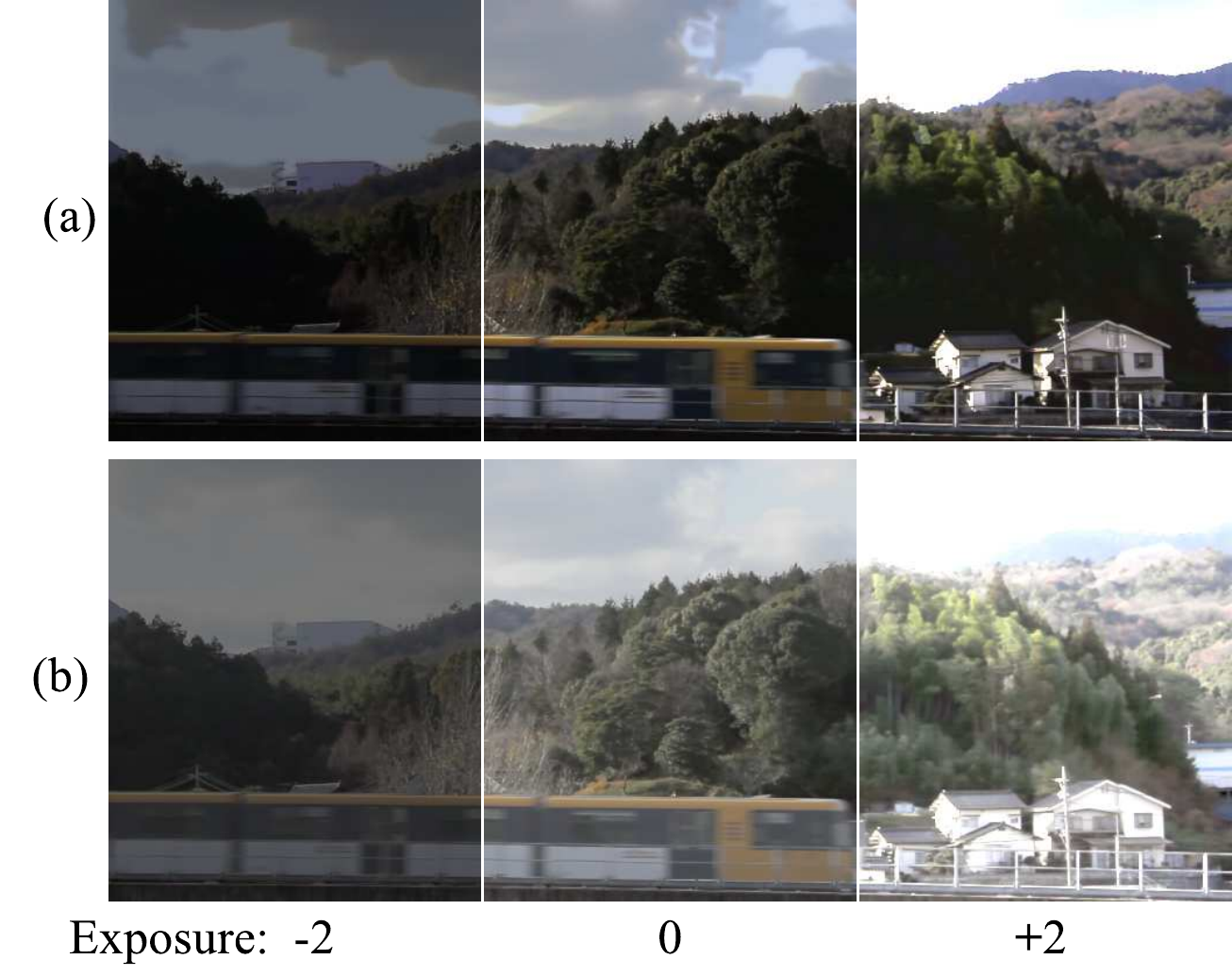}
\vspace{-3mm}
\caption{From left to right columns, a synthesized HDR image is divided into three areas, whose exposures are -2, 0, +2 respectively. (a) shows the results obtained by HDR reconstruction method from single image~\cite{Demetris18} and (b) that by our method. The gradation slightly changes on the clouds and the details of the dark forests can be retained by our method. }
\label{fig:hdrcomp}
\end{figure}

\hspace{4mm} \textbf{Motion blur on arbitrary objects}\\
Figure~\ref{fig:application}(a)
is used to apply motion blur only on 
selected objects from many moving objects in the scene. 
From the results, it is confirmed that only the target objects are sharp whereas other 
objects are all blurred out.
Please refer to the supplemental video to see the panning shot videos.

In recent years, a new method of video expression has emerged called a cinemagraph,
in which a part of the image moves and the surrounding area remains static.
The proposed method can be applied to create a cinemagraph.
In the Fig.~\ref{fig:application}(d),
only the target object moves while the surroundings remain still.
Since motion blur is added to the surrounding objects,
the video gives the viewer an impression that time is passing very slow other 
than the target object.
Please refer to the supplemental video.

\begin{figure}[t]
\centering
\includegraphics[width=0.75\linewidth, bb= 0 0 456 175]{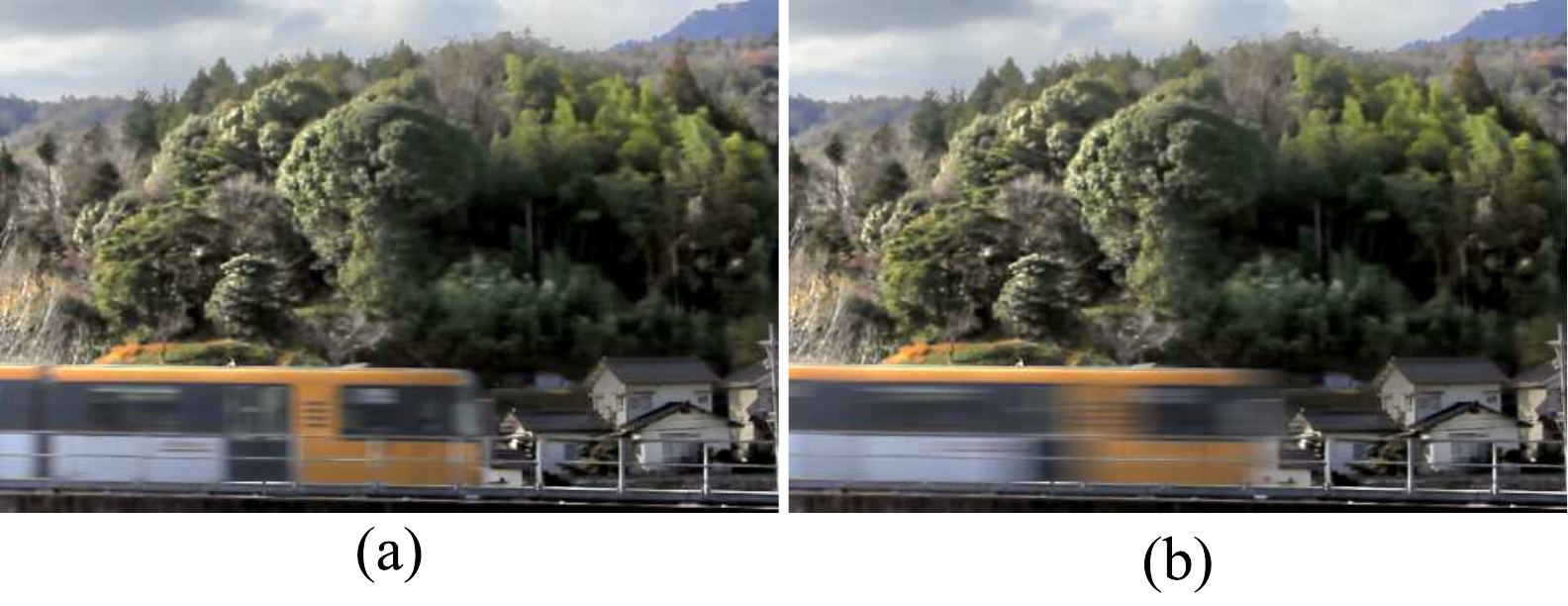}
\vspace{-5mm}
\caption{Two examples of HDR photographs including motion blur on the train. The amount of blur in (b) is emphasized by 4.5 times of (a) by the user.}
\label{fig:blurcomp}

\centering
\includegraphics[width=0.8\linewidth]{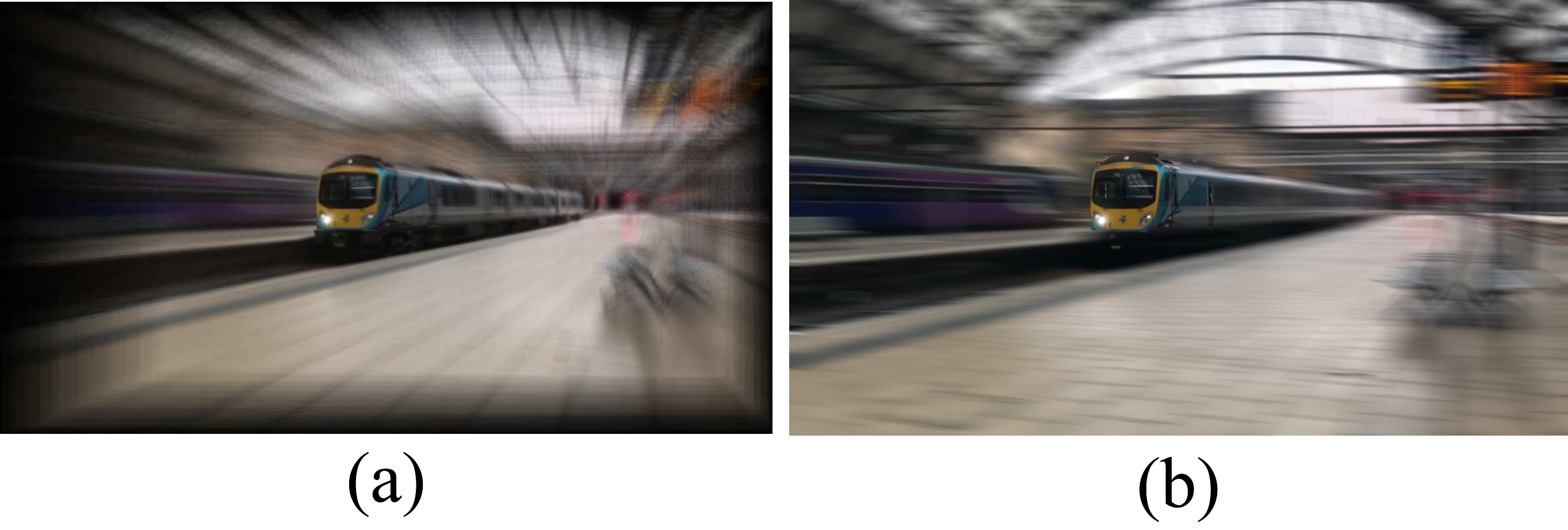}
\vspace{-5mm}
\caption{The comparison with the existing method~\cite{Lancelle19}. Our method (b) can keep the target with sharp edges more than the existing method (a).}
\label{fig:blurcomp2}
\vspace{-3mm}
\end{figure}

\textbf{The comparison of HDR image synthesis}\\
Figure~\ref{fig:hdrcomp} (a) shows results of HDR image obtained from a single frame and (b) obtained by our method using multiple exposure frames. 
From the results, it is confirmed that the HDR image keeps details in highlights 
and dark areas those are lost in a single LDR image.
In addition, our method can emphasize the dynamism of the target object by changing the size of the motion blur for the artistic purpose (Figure~\ref{fig:blurcomp}(b)).
We compare our result with the existing method~\cite{Lancelle19} in Fig.~\ref{fig:blurcomp2}.
The proposed method can keep the sharp edges of the train by using the mask image, while another uses image alignment that brings blurry edges. 

\vspace{-2mm}
\section{Conclusions}
\label{sec:Conclusions}
\vspace{-2mm}

In this paper, we proposed a method to add motion blur on arbitrary 
objects in videos or photographs.
To achieve the purpose, 
we assume that input images are blur free by capturing images with short shutter 
speed. Since short shutter speed makes images extremely dark, a large gain is 
required which causes severe noises on captured image, and thus, color 
compensation method based on nonlinear filtering is proposed.
Then, a captured images are segmented into
foreground and background layers and
a motion blur is only applied to the selected object.
Final images are synthesized by merging foreground and background images again.
HDRI is also made by capturing multiple-exposure images.
Effectiveness of the method was proved by comparison to previous method and ablation studies using real images.
Machine learning based solution to achieve robust result is our future work.

\vspace{-0.4cm}
\section*{\centering \large Acknowledgment}
\vspace{-0.3cm}
This work was supported by JSPS/KAKENHI 20H00611, 18K19824, 18H04119 in Japan.
\vspace{-0.4cm}

\clearpage
\bibliographystyle{IEEEbib}
\bibliography{references}

\end{document}